# Centroid-based summarization of multiple documents: sentence extraction, utility-based evaluation, and user studies

Dragomir R. Radev
School of Information
University of Michigan
Ann Arbor, MI 48103
radev@umich.edu

Hongyan Jing
Department of Computer Science
Columbia University
New York, NY 10027
hjing@cs.columbia.edu

Malgorzata Budzikowska
IBM TJ Watson Research Center
30 Saw Mill River Road
Hawthorne, NY 10532
sm1@us.ibm.com

## Abstract

We present a multi-document summarizer, called MEAD, which generates summaries using cluster centroids produced by a topic detection and tracking system. We also describe two new techniques, based on sentence utility and subsumption, which we have applied to the evaluation of both single and multiple document summaries. Finally, we describe two user studies that test our models of multi-document summarization.

## 1 Introduction

On October 12, 1999, a relatively small number of news sources mentioned in passing that Pakistani Defense Minister Gen. Pervaiz Musharraf was away visiting Sri Lanka. However, all world agencies would be actively reporting on the major events that were to happen in Pakistan in the following days: Prime Minister Nawaz Sharif announced that in Gen. Musharraf's absence, the Defense Minister had been sacked and replaced by General Zia Addin. Large numbers of messages from various sources started to inundate the newswire: about the army's occupation of the capital, the Prime Minister's ouster and his subsequent placement under house arrest, Gen. Musharraf's return to his country, his ascendancy to power, and the imposition of military control over Pakistan.

The paragraph above summarizes a large amount of news from different sources. While it was not automatically generated, one can imagine the use of such automatically generated summaries. In this paper we will describe how multi-document summaries are built and evaluated.

### 1.1 Topic detection and multi-document summarization

The process of identifying all articles on an emerging event is called *Topic Detection and Tracking (TDT)*. A large body of research in TDT has been created over the past two years [Allan et al., 98]. We will present an extension of our own research on TDT [Radev et al., 1999] to cover summarization of multi-document clusters.

Our entry in the official TDT evaluation, called CIDR [Radev et al., 1999], uses modified TF*IDF to produce clusters of news articles on the same event. We developed a new technique for multi-document summarization (or MDS), called *centroid-based summarization (CBS)* which uses as input the centroids of the clusters produced by CIDR to identify which sentences are central to the topic of the cluster, rather than the individual articles. We have implemented CBS in a system, named MEAD.

The main contributions of this paper are: the development of a centroid-based multi-document summarizer, the use of *cluster-based sentence utility* (CBSU) and *cross-sentence informational subsumption* (CSIS) for evaluation of single and multi-document summaries, two user studies that support our findings, and an evaluation of MEAD.

An *event cluster,* produced by a TDT system, consists of chronologically ordered news articles from multiple sources, which describe an event as it develops over time. Event clusters range from 2 to 10 documents from which MEAD produces summaries in the form of sentence extracts.

A key feature of MEAD is its use of cluster centroids, which consist of words which are central not only to *one* article in a cluster, but to *all* the articles.

MEAD is significantly different from previous work on multi-document summarization [Radev & McKeown, 1998; Carbonell and Goldstein, 1998; Mani and Bloedorn, 1999; McKeown et al., 1999], which use techniques such as graph matching, maximal marginal relevance, or language generation.

Finally, evaluation of multi-document summaries is a difficult problem. There is not yet a widely accepted evaluation scheme. We propose a utility-based evaluation scheme, which can be used to evaluate both single-document and multi-document summaries.

## 2 Informational content of sentences

### 2.1 Cluster-based sentence utility (CBSU)

*Cluster-based sentence utility* (CBSU, or utility) refers to the degree of relevance (from 0 to 10) of a particular sentence to the general topic of the entire cluster (for a discussion of what is a topic, see [Allan et al. 1998]). A utility of 0 means that the sentence is not relevant to the cluster and a 10 marks an essential sentence.

### 2.2 Cross-sentence informational subsumption (CSIS)

A related notion to CBSU is *cross-sentence informational subsumption* (CSIS, or subsumption), which reflects that certain sentences repeat some of the information present in other sentences and may, therefore, be omitted during summarization. If the information content of sentence **a** (denoted as **i(a)**) is contained within sentence **b,** then **a** becomes informationally redundant and the content of **b** is said to *subsume* that of **a**:

$$i(a) \subset i(b)$$

In the example below, (2) subsumes (1) because the crucial information in (1) is also included in (2) which presents additional content: "the court", "last August", and "sentenced him to life".

(1) John Doe was found guilty of the murder.
(2) The court found John Doe guilty of the murder of Jane Doe last August and sentenced him to life.

The cluster shown in Figure 1 shows subsumption links across two articles[1] about recent terrorist activities in Algeria (ALG 18853 and ALG 18854).

An arrow from sentence A to sentence B indicates that the information content of A is subsumed by the information content of B. Sentences 2, 4, and 5 from the first article repeat the information from sentence 2 in the second article, while sentence 9 from the former article is later repeated in sentences 3 and 4 of the latter article.

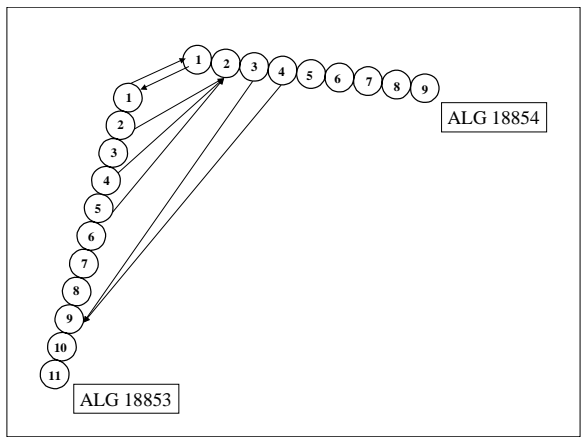

**Figure 1: Subsumption links across two articles: ALG 18853 and ALG 18854.**

### 2.3 Equivalence classes of sentences

Sentences subsuming each other are said to belong to the same *equivalence class*. An equivalence class may contain more than two sentences within the same or different articles. In the following example, although sentences (3) and (4) are not exact paraphrases of each other, they can be substituted for each other without crucial loss of information and therefore belong to the same equivalence class, i.e. $i(3) \subset i(4)$ and $i(4) \subset i(3)$. In the user study section we will take a look at the way humans perceive CSIS and equivalence class.

(3) Eighteen decapitated bodies have been found in a mass grave in northern Algeria, press reports said Thursday.
(4) Algerian newspapers have reported on Thursday that 18 decapitated bodies have been found by the authorities.

### 2.4 Comparison with MMR

Maximal marginal relevance (or MMR) is a technique similar to CSIS and was introduced in [Carbonell and Goldstein, 1998]. In that paper, MMR is used to produce summaries of single documents that avoid redundancy. The authors mention that their preliminary results indicate that multiple documents on the same topic also contain redundancy but they fall short of using MMR for multi-document summarization. Their metric is used as an enhancement to a query-based summary whereas CSIS is designed for query-independent (a.k.a., generic) summaries.

---
[1] The full text of these articles is shown in the Appendix.

## 3 MEAD: a centroid-based multi-document summarizer

We now describe the corpus used for the evaluation of MEAD, and later in this section we present MEAD's algorithm.

| Cluster | # docs | # sent | source | news sources | topic |
|---------|--------|--------|--------|--------------|-------|
| A | 2 | 25 | clari.world.africa.northwestern | AFP, UPI | Algerian terrorists threaten Belgium |
| B | 3 | 45 | clari.world.terrorism | AFP, UPI | The FBI puts Osama bin Laden on the most wanted list |
| C | 2 | 65 | clari.world.europe.russia | AP, AFP | Explosion in a Moscow apartment building (September 9, 1999) |
| D | 7 | 189 | clari.world.europe.russia | AP, AFP, UPI | Explosion in a Moscow apartment building (September 13, 1999) |
| E | 10 | 151 | TDT-3 corpus topic 78 | AP, PRI, VOA | General strike in Denmark |
| F | 3 | 83 | TDT-3 corpus topic 67 | AP, NYT | Toxic spill in Spain |

**Table 1: Corpus composition**

### 3.1 Description of the corpus

For our experiments, we prepared a small corpus consisting of a total of 558 sentences in 27 documents, organized in 6 clusters (Table 1), all extracted by CIDR. Four of the clusters are from Usenet newsgroups. The remaining two clusters are from the official TDT corpus[2]. Among the factors for our selection of clusters are: coverage of as many news sources as possible, coverage of both TDT and non-TDT data, coverage of different types of news (e.g., terrorism, internal affairs, and environment), and diversity in cluster sizes (in our case, from 2 to 10 articles). The test corpus is used in the evaluation in such a way that each cluster is summarized at 9 different compression rates, thus giving nine times as many sample points as one would expect from the size of the corpus.

### 3.2 Cluster centroids

Table 2 shows a sample centroid, produced by CIDR [Radev et al., 1999] from cluster A. The "count" column indicates the average number of occurrences of a word across the entire cluster. The IDF values were computed from the TDT corpus. A centroid, in this context, is a pseudo-document which consists of words which have Count*IDF scores above a pre-defined threshold in the documents that constitute the cluster. CIDR computes Count*IDF in an iterative fashion, updating its values as more articles are inserted in a given cluster. We hypothesize that sentences that contain the words from the centroid are more indicative of the topic of the cluster.

| Word | Count | IDF | Count * IDF |
|------|-------|-----|-------------|
| belgium | 15.50 | 4.96 | 76.86 |
| gia | 7.50 | 8.39 | 62.90 |
| algerian | 6.00 | 6.36 | 38.15 |
| hayat | 3.00 | 8.90 | 26.69 |
| algeria | 4.50 | 5.63 | 25.32 |
| islamic | 6.00 | 4.13 | 24.76 |
| melouk | 2.00 | 10.00 | 19.99 |
| arabic | 3.00 | 5.99 | 17.97 |
| battalion | 2.50 | 7.16 | 17.91 |

**Table 2: Sample centroid produced by CIDR**

### 3.3 Centroid-based algorithm

MEAD decides which sentences to include in the extract by ranking them according to a set of parameters. The input to MEAD is a cluster of articles (e.g., extracted by CIDR) and a value for the compression rate $r$. For example, if the cluster contains a total of 50 sentences ($n = 50$) and the value of $r$ is 20%, the output of MEAD will contain 10 sentences. Sentences are laid in the same order as they appear in the original documents with documents ordered chronologically. We benefit here from the time stamps associated with each document.

$$SCORE\ (s) = \sum_i (w_c C_i + w_p P_i + w_f F_i)$$

where $i\ (1 \leq i \leq n)$ is the sentence number within the cluster.

INPUT: Cluster of $d$ documents[3] with $n$ sentences (compression rate $= r$)

---

[2] The selection of Cluster E is due to an idea by the participants in the Novelty Detection Workshop, led by James Allan.

[3] Note that currently, MEAD requires that sentence boundaries be marked.

OUTPUT: (n * r) sentences from the cluster with the highest values of SCORE.

The current paper evaluates two algorithms (pure centroid: $w_c = 1$, $w_p = 0$, $w_f = 0$) and (lead+centroid: $w_c = 1$, $w_p = 1$, $w_f = 0$).

### 3.4 Redundancy-based algorithm

We try to approximate CSIS by identifying sentence similarity across sentences. Its effect on MEAD is the subtraction of a *redundancy penalty* ($R_s$) for each sentence which overlaps with sentences that have higher SCORE values. The redundancy penalty is similar to the negative factor in the MMR formula [Carbonell and Goldstein, 1998].

$$SCORE(s) = \sum_i (w_c C_i + w_p P_i + w_f F_i) - w_R R_s$$

For each pair of sentences extracted by MEAD, we compute the *cross-sentence word overlap* according to the following formula:

$R_s$ = 2 * (# overlapping words) / (# words in sentence 1 + # words in sentence 2)

$$w_R = Max_s(SCORE(s))$$

$R_s = 1$ when the sentences are identical and $R_s = 0$ when they have no words in common. After deducting $R_s$, we rerank all sentences and possibly create a new sentence extract. We repeat this process until reranking doesn't result in a different extract.

The number of overlapping words in the formula is computed in such a way that if a word appears $m$ times in one sentence and $n$ times in another, only *min (m, n)* of these occurrences will be considered overlapping.

## 4 Techniques for evaluating summaries

Summarization evaluation methods can be divided into 2 categories: intrinsic and extrinsic [Mani and Maybury, 1999]. Intrinsic evaluation measures the quality of summaries directly (e.g., by comparing them to ideal summaries). Extrinsic methods measure how well the summaries help in performing a particular task (e.g., classification). The extrinsic evaluation, also called task-based evaluation, has received more attention recently at the DARPA Summarization Evaluation Conference [Mani et al., 1998].

### 4.1 Single-document summaries

Two techniques commonly used to measure interjudge agreement and to evaluate extracts are (A), precision and recall, and (B), percent agreement. In both cases, an automatically generated summary is compared against an "ideal" summary. To construct the ideal summary, a group of human subjects are asked to extract sentences. Then, the sentences chosen by a majority of humans are included in the ideal summary. The precision and recall indicate the overlap between the ideal summary and the automatic summary.

We should note that [Jing et al., 1998] pointed out that the cut-off summary length can affect results significantly, and the assumption of a single "ideal" summary is problematic.

We will illustrate why these two methods are not satisfactory. Suppose we want to determine which of two systems which selected summary sentences at a compression rate of 20% (Table 3) is better.

|     | Ideal | System1 | System2 |
| --- | --- | --- | --- |
| S1  | +     | +       | -       |
| S2  | +     | -       | -       |
| S3  | -     | +       | +       |
| S4  | -     | -       | +       |
| S5  | -     | -       | -       |
| S6  | -     | -       | -       |
| S7  | -     | -       | -       |
| S8  | -     | -       | -       |
| S9  | -     | -       | -       |
| S10 | -     | -       | -       |

**Table 3: Comparing systems without utility metrics**

Using precision and recall indicates that the performance of System 1 and System 2 is 50% and 0%, respectively. System 2 appears to perform in the worst possible way since it is not possible to differentiate between sentences S3 – S10, which are equally bad in this model. Using percent agreement, the performance is 80% and 60%, respectively, however percent agreement is highly dependent on the compression rate.

### 4.2 Utility-based evaluation of both single and multiple document summaries.

Instead of P&R or percent agreement, one can measure the coverage of the ideal summary's utility. In the example in Table 4, using both evaluation methods A and B, System 1 achieves 50%, whereas System 2 achieves 0%. If we look at sentence utility, System 1 matches 18 out of 19 utility points in the ideal summary and System 2 gets 15 out of 19. In this case, the performance of system 2 is not as low as when using methods A and B.

|    | *Ideal* | *System 1* | *System 2* |
|----|---------|------------|------------|
| S1 | 10 (+)  | +          | -          |
| S2 | 9 (+)   | -          | -          |
| S3 | 8       | +          | +          |
| S4 | 7       | -          | +          |

**Table 4: Comparing systems with utility metrics**

We therefore propose to model both interjudge agreement and system evaluation as real-valued vector matching and not as boolean (methods A and B). By giving credit for "less than ideal" sentences and distinguishing the degree of importance between sentences, the utility-based scheme is a more natural model to evaluate summaries.

Other researchers have also suggested improvements on the precision and recall measure for summarization. [Jing et al., 1998] proposed to use fractional P&R. [Goldstein et al., 1999] used 11-point average precision.

### 4.2.1 Interjudge agreement (J)

Without loss of generality, suppose that three judges are asked to build extracts of a single article[4]. As an example, Table 5 shows the weights of the different sentences (note that no compression rate needs to be specified; from the data in the table, one can generate summaries at arbitrary compression rates).

|            | *Judge1* | *Judge2* | *Judge3* |
|------------|----------|----------|----------|
| Sentence 1 | 10       | 10       | 5        |
| Sentence 2 | 8        | 9        | 8        |
| Sentence 3 | 2        | 3        | 4        |
| Sentence 4 | 5        | 6        | 9        |

**Table 5: Illustrative example**

The interjudge agreement measures, to what extent each judge satisfies the utility of the other judges by picking the right sentences.

In the example, with a 50% summary, Judge 1 would pick sentences 1 and 2 because they have the maximum utility as far as he is concerned. Judge 2 would select the same two sentences, while Judge 3 would pick 2 and 4[5]. The maximum utilities for each judge are as follows: 18 (= 10 + 8), 19, and 17.

How well Judge 1's utility assignment satisfies Judge 2's utility need? Since they have both selected the same sentences, Judge 1 achieves 19/19 (1.00) of Judge 2's utility. However, Judge 1 only achieves 13/17 (0.765) of Judge 3's utility.

We can therefore represent the cross-judge utility agreement $J_{i,j}$ as an asymmetric matrix (e.g., the value of $J_{1,2}$ is 0.765 while the value of $J_{2,1}$ is 13/18 or 0.722). The values $J_{i,j}$ of the cross-judge utility matrix for r = 50% are shown in Table 6.

|         | *Judge 1* | *Judge 2* | *Judge 3* | *Overall* |
|---------|-----------|-----------|-----------|-----------|
| Judge 1 | 1.000     | 1.000     | 0.765     | 0.883     |
| Judge 2 | 1.000     | 1.000     | 0.765     | 0.883     |
| Judge 3 | 0.722     | 0.789     | 1.000     | 0.756     |

**Table 6: Cross-judge utility agreement (J)**

We can also compute the performance of each judge ($J_i$) against all other judges by averaging for each Judge i all values in the matrix $J_{i,j}$ where $i \neq j$. These numbers indicate that Judge 3 is the outlier.

Finally, the mean cross-judge agreement J is the average of $J_i$ for i=1..3. In the example, J = 0.841.

J is like an upper bound on the performance of a summarizer (it can achieve a score higher than J only when it can do a better job than the judges).

### 4.2.2 Random performance (R)

We can also similarly define a lower bound on the summarizer performance.

The random performance R is the average of all possible system outputs at a given compression rate, *r*. For example, with 4 sentences and a *r* = 50%, the set of all possible system outputs is {12,13,14,23,24,34}[6]. For each of them, we can compute a system performance. For example, the system that selects sentences 1 and 4 (we label this system as {14}) performs at 15/18 (or 0.833) against Judge 1, at 16/19 against Judge 2 (or 0.842), and at 14/17 against Judge 3 (or 0.824). On average, the performance of {14} is the average of the three numbers, or 0.833.

We can compute the performance of *all* possible systems. The six numbers (in the order {12,13,14,23,24,34}) are 0.922, 0.627, 0.833, 0.631, 0.837, and 0.543. Their average becomes the random performance (R) of all possible systems; in this example, R = 0.732.

---

[4] We concatenate all documents in a cluster in a chronological order.

[5] In case of ties, we arbitrarily pick the sentence that occurs earlier in the cluster.

[6] There are a total of (n!) / (n(1-r))! (r*n)! system outputs.

### 4.2.3 System performance (S)

The system performance S is one of the numbers[6] described in the previous subsection. For {13}, the value of S is 0.627 (which is lower than random). For {14}, S is 0.833, which is between R and J. In the example, only two of the six possible sentence selections, {14} and {24} are between R and J. Three others, {13}, {23}, and {34} are below R. while {12} is better than J.

### 4.2.4. Normalized system performance (D)

To restrict system performance (mostly) between 0 and 1, we use a mapping between R and J in such a way that when S = R, the normalized system performance, D, is equal to 0 and when S = J, D becomes 1. The corresponding linear function[7] is:

$$D = (S-R) / (J-R)$$

Figure 2 shows the mapping between system performance S on the left (a) and normalized system performance D on the right (b). A small part of the 0-1 segment is mapped to the entire 0-1 segment; therefore the difference between two systems, performing at e.g., 0.785 and 0.812 can be significant!

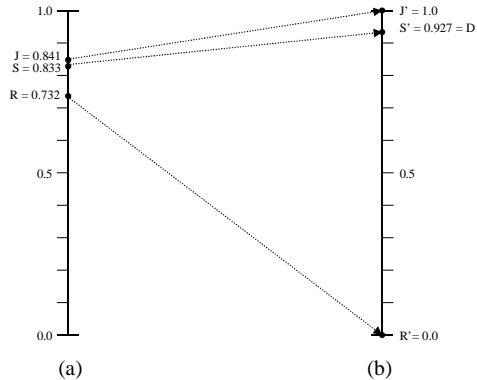

**Figure 2: Performance mapping**

Example: the normalized system performance for the {14} system then becomes (0.833 - 0.732) / (0.841 – 0.732) or 0.927. Since the score is close to 1, the {14} system is almost as good as the interjudge agreement. The normalized system performance for the {24} system is similarly (0.837 – 0.732) / (0.841 – 0.732) or 0.963. Of the two systems, {24} outperforms {14}.

## 4.3 Using CSIS to evaluate multi-document summaries

To use CSIS in the evaluation, we introduce a new parameter, E, which tells us how much to penalize a system that includes redundant information. In the example from Table 7 (arrows indicate subsumption), a summarizer with $r = 20\%$ needs to pick 2 out of 12 sentences. Suppose that it picks 1/1 and 2/1 (in bold). If E = 1, it should get full credit of 20 utility points. If E = 0, it should get no credit for the second sentence as it is subsumed by the first sentence. By varying E between 0 and 1, the evaluation may favor or ignore subsumption.

|  | *Article1* | *Article2* | *Article3* |
|---|---|---|---|
| Sent1 | **10** ⟶ | **10** | 5 |
| Sent2 | 8 | 9 | 8 |
| Sent3 | 2 | 3 | 4 |
| Sent4 | 5 | 6 | 9 |

**Table 7: Sample subsumption table (12 sentences, 3 articles)**

## 5 User studies and system evaluation

We ran two user experiments. First, six judges were each given six clusters and asked to ascribe an importance score from 0 to 10 to each sentence within a particular cluster. Next, five judges had to indicate for each sentence which other sentence(s), if any, it subsumes[8].

### 5.1 CBSU: interjudge agreement

Using the techniques described in Section 0, we computed the cross-judge agreement (J) for the 6 clusters for various $r$ (Figure 3). Overall, interjudge agreement was quite high. An interesting drop in interjudge agreement occurs for 20-30% summaries. The drop most likely results from the fact that 10% summaries are typically easier to produce because the few most important sentences in a cluster are easier to identify.

---

[7] The formula is valid when J > R (that is, the judges agree among each other better than randomly).

[8] We should note that both annotation tasks were quite time consuming and frustrating for the users who took anywhere from 6 to 10 hours each to complete their part.

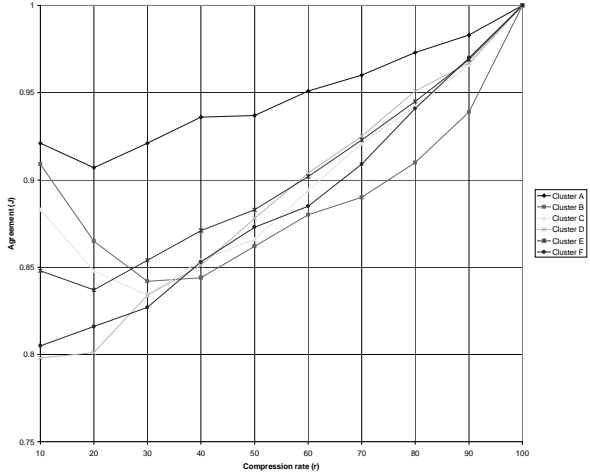

**Figure 3: Cross-judge agreement (J) on the CBSU annotation task.**

## 5.2 CSIS: interjudge agreement

In the second experiment, we asked users to indicate all cases when within a cluster, a sentence is subsumed by another. The judges' data on the first seven sentences of cluster A are shown in Table 8.

The "+ score" indicates the number of judges who agree on the most frequent subsumption. The "- score" indicates that the consensus was no subsumption. We found relatively low interjudge agreement on the cases in which at least one judge indicated evidence of subsumption. Overall, out of 558 sentences, there was full agreement (5 judges) on 292 sentences (Table 9). Unfortunately, in 291 of these 292 sentences the agreement was that there is no subsumption. When the bar of agreement was lowered to four judges, 23 out of 406 agreements are on sentences with subsumption. Overall, out of 80 sentences with subsumption, only 24 had an agreement of four or more judges. However, in 54 cases at least three judges agreed on the presence of a particular instance of subsumption.

| Sentence | Judge1 | Judge2 | Judge3 | Judge4 | Judge5 | + score | - score |
|---|---|---|---|---|---|---|---|
| A1-1 | - | A2-1 | A2-1 | - | A2-1 | 3 | |
| A1-2 | A2-5 | A2-5 | - | - | A2-5 | 3 | |
| A1-3 | - | - | - | - | A2-10 | | 4 |
| A1-4 | A2-10 | A2-10 | A2-10 | - | A2-10 | 4 | |
| A1-5 | - | A2-1 | - | A2-2 | A2-4 | | 2 |
| A1-6 | - | - | - | - | A2-7 | | 4 |
| A1-7 | - | - | - | - | A2-8 | | 4 |

**Table 8: Judges' indication for subsumption for the first seven sentences in cluster A**

| | Cluster A | | Cluster B | | Cluster C | | Cluster D | | Cluster E | | Cluster F | |
|---|---|---|---|---|---|---|---|---|---|---|---|---|
| # judges agreeing | + | - | + | - | + | - | + | - | + | - | + | - |
| 5 | 0 | 7 | 0 | 24 | 0 | 45 | 0 | 88 | 1 | 73 | 0 | 61 |
| 4 | 1 | 6 | 3 | 6 | 1 | 10 | 9 | 37 | 8 | 35 | 0 | 11 |
| 3 | 3 | 6 | 4 | 5 | 4 | 4 | 28 | 20 | 5 | 23 | 3 | 7 |
| 2 | 1 | 1 | 2 | 1 | 1 | 0 | 7 | 0 | 7 | 0 | 1 | 0 |

**Table 9: Interjudge CSIS agreement**

In conclusion, we found very high interjudge agreement in the first experiment and moderately low agreement in the second experiment. We concede that the time necessary to do a proper job at the second task is partly to blame.

## 5.3 Evaluation of MEAD

Since the baseline of random sentence selection is already included in the evaluation formulae, we used the Lead-based method (selecting the positionally first (n*r/c) sentences from each cluster where c = number of clusters) as the baseline to evaluate our system.

In Table 10 we show the *normalized* performance (D) of MEAD, for the six clusters at nine compression rates. MEAD performed better than Lead in 29 (in bold) out of 54 cases. Note that for the largest cluster, Cluster D, MEAD outperformed Lead at all compression rates.

|           | *10%* | *20%* | *30%* | *40%* | *50%* | *60%* | *70%* | *80%* | *90%* |
|-----------|-------|-------|-------|-------|-------|-------|-------|-------|-------|
| Cluster A | 0.855 | 0.572 | 0.427 | **0.759** | **0.862** | **0.910** | 0.554 | **1.001** | 0.584 |
| Cluster B | 0.365 | 0.402 | 0.690 | 0.714 | **0.867** | 0.640 | 0.845 | 0.713 | 1.317 |
| Cluster C | 0.753 | **0.938** | 0.841 | **1.029** | 0.751 | 0.819 | 0.595 | 0.611 | 0.683 |
| Cluster D | **0.739** | 0.764 | 0.683 | 0.723 | 0.614 | 0.568 | 0.668 | 0.719 | **1.100** |
| Cluster E | **1.083** | 0.937 | 0.581 | 0.373 | 0.438 | 0.369 | 0.429 | 0.487 | 0.261 |
| Cluster F | 1.064 | 0.893 | **0.928** | **1.000** | 0.732 | **0.805** | 0.910 | 0.689 | 0.199 |

**Table 10: Normalized performance (D) of MEAD**

We then modified the MEAD algorithm to include lead information as well as centroids (see Section 0). In this case, MEAD+Lead performed better than the Lead baseline in 41 cases. We are in the process of running experiments with other SCORE formulas.

### 5.4 Discussion

It may seem that utility-based evaluation requires too much effort and is prone to low interjudge agreement. We believe that our results show that interjudge agreement is quite high. As far as the amount of effort required, we believe that the larger effort on the part of the judges is more or less compensated with the ability to evaluate summaries off-line and at variable compression rates. Alternative evaluations don't make such evaluations possible. We should concede that a utility-based approach is probably not feasible for query-based summaries as these are typically done only on-line.

We discussed the possibility of a sentence contributing negatively to the utility of another sentence due to redundancy. We should also point out that sentences can also reinforce one another positively. For example, if a sentence mentioning a new entity is included in a summary, one might also want to include a sentence that puts the entity in the context of the rest of the article or cluster.

## 6 Contributions and future work

We presented a new multi-document summarizer, MEAD. It summarizes clusters of news articles automatically grouped by a topic detection system. MEAD uses information from the centroids of the clusters to select sentences that are most likely to be relevant to the cluster topic.

We used a new utility-based technique, CBSU, for the evaluation of MEAD and of summarizers in general. We found that MEAD produces summaries that are similar in quality to the ones produced by humans. We also compared MEAD's performance to an alternative method, multi-document lead, and showed how MEAD's sentence scoring weights can be modified to produce summaries significantly better than the alternatives.

We also looked at a property of multi-document clusters, namely cross-sentence information subsumption (which is related to the MMR metric proposed in [Carbonell and Goldstein, 1998]) and showed how it can be used in evaluating multi-document summaries.

All our findings are backed by the analysis of two experiments that we performed with human subjects. We found that the interjudge agreement on sentence utility is very high while the agreement on cross-sentence subsumption is moderately low, although promising.

In the future, we would like to test our multidocument summarizer on a larger corpus and improve the summarization algorithm. We would also like to explore how the techniques we proposed here can be used for multiligual multidocument summarization.

## 7 Acknowledgments

We would like to thank Inderjeet Mani, Wlodek Zadrozny, Rie Kubota Ando, Joyce Chai, and Nanda Kambhatla for their valuable feedback. We would also like to thank Carl Sable, Min-Yen Kan, Dave Evans, Adam Budzikowski, and Veronika Horvath for their help with the evaluation.

# Appendix

**ARTICLE 18853:** ALGIERS, May 20 (AFP)

1. Eighteen decapitated bodies have been found in a mass grave in northern Algeria, press reports said Thursday, adding that two shepherds were murdered earlier this week.

2. Security forces found the mass grave on Wednesday at Chbika, near Djelfa, 275 kilometers (170 miles) south of the capital.

3. It contained the bodies of people killed last year during a wedding ceremony, according to Le Quotidien Liberte.

4. The victims included women, children and old men.

5. Most of them had been decapitated and their heads thrown on a road, reported the Es Sahafa.

6. Another mass grave containing the bodies of around 10 people was discovered recently near Algiers, in the Eucalyptus district.

7. The two shepherds were killed Monday evening by a group of nine armed Islamists near the Moulay Slissen forest.

8. After being injured in a hail of automatic weapons fire, the pair were finished off with machete blows before being decapitated, Le Quotidien d'Oran reported.

9. Seven people, six of them children, were killed and two injured Wednesday by armed Islamists near Medea, 120 kilometers (75 miles) south of Algiers, security forces said.

10. The same day a parcel bomb explosion injured 17 people in Algiers itself.

11. Since early March, violence linked to armed Islamists has claimed more than 500 lives, according to press tallies.

**ARTICLE 18854:** ALGIERS, May 20 (UPI)

1. Algerian newspapers have reported that 18 decapitated bodies have been found by authorities in the south of the country.

2. Police found the "decapitated bodies of women, children and old men, with their heads thrown on a road" near the town of Jelfa, 275 kilometers (170 miles) south of the capital Algiers.

3. In another incident on Wednesday, seven people -- including six children -- were killed by terrorists, Algerian security forces said.

4. Extremist Muslim militants were responsible for the slaughter of the seven people in the province of Medea, 120 kilometers (74 miles) south of Algiers.

5. The killers also kidnapped three girls during the same attack, authorities said, and one of the girls was found wounded on a nearby road.

6. Meanwhile, the Algerian daily Le Matin today quoted Interior Minister Abdul Malik Silal as saying that "terrorism has not been eradicated, but the movement of the terrorists has significantly declined."

7. Algerian violence has claimed the lives of more than 70,000 people since the army cancelled the 1992 general elections that Islamic parties were likely to win.

8. Mainstream Islamic groups, most of which are banned in the country, insist their members are not responsible for the violence against civilians.

9. Some Muslim groups have blamed the army, while others accuse "foreign elements conspiring against Algeria."

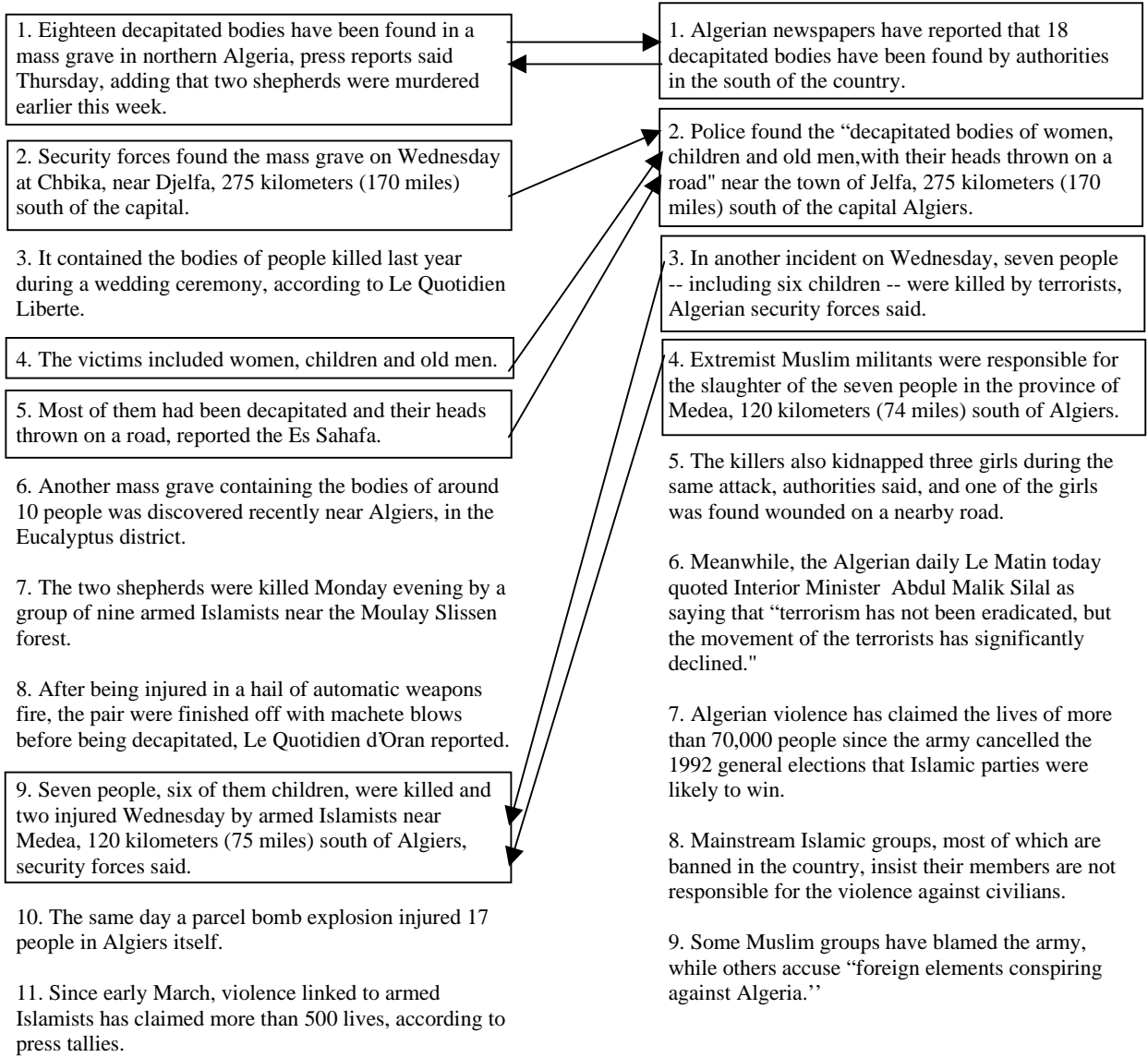